\documentclass[letterpaper]{article} 
\usepackage{aaai2026}  
\usepackage{times}  
\usepackage{helvet}  
\usepackage{courier}  
\usepackage[hyphens]{url}  
\usepackage{graphicx} 
\usepackage{natbib}  
\usepackage{caption} 
\usepackage{algorithm}
\usepackage{algorithmic}

\usepackage{newfloat}
\usepackage{listings}
\DeclareCaptionStyle{ruled}{labelfont=normalfont,labelsep=colon,strut=off} 
\floatstyle{ruled}
\newfloat{listing}{tb}{lst}{}
\floatname{listing}{Listing}
\usepackage{booktabs}
\usepackage[table]{xcolor}
\usepackage{array}
\usepackage{amssymb}
\usepackage{amsmath}
\usepackage{pifont}
\usepackage{multirow}
\usepackage{subcaption}
\usepackage{times}
\usepackage{xcolor}
\usepackage{makecell}

\title{InfoCLIP: Bridging Vision-Language Pretraining and Open-Vocabulary \\ Semantic Segmentation via Information-Theoretic Alignment Transfer}
\author{
    Muyao Yuan\textsuperscript{\rm 1,2}, Yuanhong Zhang\textsuperscript{\rm 1,3}, Weizhan Zhang\textsuperscript{\rm 1,2}\thanks{Corresponding authors.}, \\
    Lan Ma\textsuperscript{\rm 4}, Yuan Gao\textsuperscript{\rm 4}, Jiangyong Ying\textsuperscript{\rm 5}, Yudeng Xin\textsuperscript{\rm 6}
}
\affiliations{
    \textsuperscript{\rm 1}School of Computer Science and Technology, Xi’an Jiaotong University\\
    \textsuperscript{\rm 2}Ministry of Education Key Laboratory of Intelligent Networks and Network Security, Xi’an Jiaotong University\\
    \textsuperscript{\rm 3}Shaanxi Province Key Laboratory of Big Data Knowledge Engineering, Xi’an Jiaotong University\\
    \textsuperscript{\rm 4}China Telecom\\
    \textsuperscript{\rm 5}China Telecom E-surfing Vision Technology Co., Ltd \\
    \textsuperscript{\rm 6}Faculty of Engineering and Information Technology, University of Melbourne

    \{yuanmuyao,yuanhongzhang\}@stu.xjtu.edu.cn, zhangwzh@xjtu.edu.cn, \\
    \{malan,gaoy97,yingjiangyong\}@chinatelecom.cn, xinyudeng005@gmail.com
}

\usepackage{bibentry}

\begin{document}

\maketitle

\begin{abstract}
Recently, the strong generalization ability of CLIP has facilitated open-vocabulary semantic segmentation, which labels pixels using arbitrary text.
However, existing methods that fine-tune CLIP for segmentation on limited seen categories often lead to overfitting and degrade the pretrained vision-language alignment.
To stabilize modality alignment during fine-tuning, we propose InfoCLIP, which leverages an information-theoretic perspective to transfer alignment knowledge from pretrained CLIP to the segmentation task. Specifically, this transfer is guided by two novel objectives grounded in mutual information.
First, we compress the pixel-text modality alignment from pretrained CLIP to reduce noise arising from its coarse-grained local semantic representations learned under image-text supervision. Second, we maximize the mutual information between the alignment knowledge of pretrained CLIP and the fine-tuned model to transfer compact local semantic relations suited for the segmentation task.
Extensive evaluations across various benchmarks validate the effectiveness of InfoCLIP in enhancing CLIP fine-tuning for open-vocabulary semantic segmentation, demonstrating its adaptability and superiority in asymmetric transfer.
\end{abstract}

\begin{links}
    \link{Project}{https://muyaoyuan.github.io/InfoCLIP-Page}
\end{links}

\section{Introduction}
Open-vocabulary semantic segmentation (OVSS) aims to build a model that assigns pixel-level semantic labels based on arbitrary text descriptions, without being restricted to a closed category set. Vision-language foundation models~\cite{radford2021clip,jia2021align,li2022blip}, particularly CLIP~\cite{radford2021clip}, are commonly fine-tuned to enable such open-vocabulary recognition. This fine-tuning process primarily serves to transfer a vision-language model pretrained on image-text alignment to the pixel-level prediction task~\cite{cho2024catseg,jiao2024maftplus}.

\begin{figure}[ht]
\centering
\includegraphics[width=0.47\textwidth]{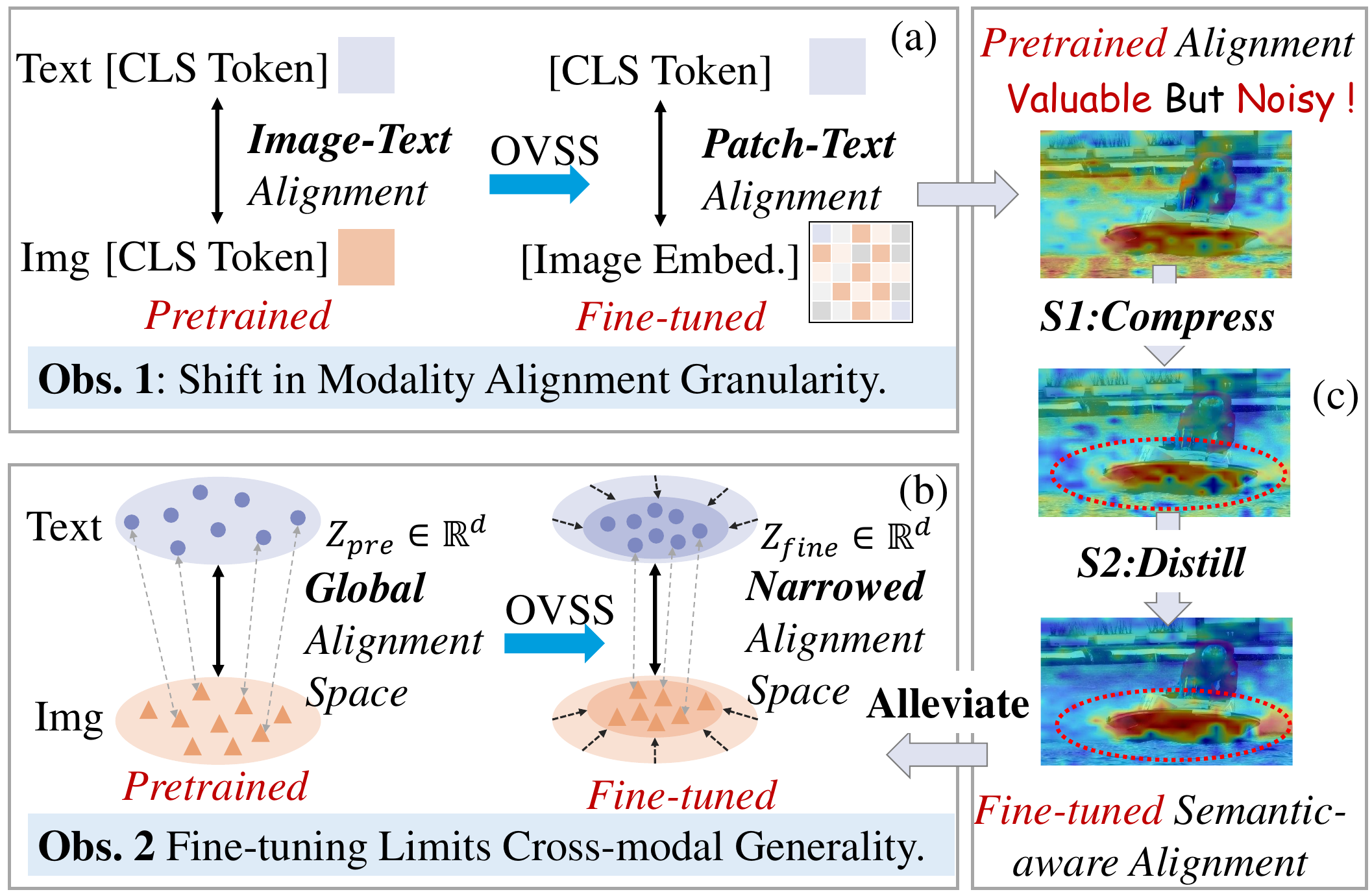}
\caption{\textbf{Motivation of InfoCLIP.} To leverage the valuable yet noisy pixel-text modality alignment from the pretrained CLIP for enhancing OVSS, we: (1) denoise the pretrained alignment through compression to extract semantic-aware alignment; and (2) transfer more generalized semantic alignment via distillation to alleviate the narrowing of the modality alignment space.}
\label{fig:motivation}
\end{figure}

Existing methods~\cite{xu2023san,jiao2023maft,cho2024catseg,jiao2024maftplus} attempt to fine-tune CLIP on a benchmark dataset for open-vocabulary segmentation. As illustrated in Fig.~\ref{fig:motivation}(b), task-specific fine-tuning on limited training categories often impairs generalization by narrowing the modality alignment space. Most methods~\cite{jiao2023maft,jiao2024maftplus,cho2024catseg} strive to limit alignment shift by simply freezing model parameters, which serves as an implicit preservation of the pretraining. However, this strategy is fragile~\cite{cho2024catseg}, as modifying even a small subset of parameters can distort the feature distribution and further compromise the image-text alignment.

To preserve the pretrained modality alignment for better generalization, an intuitive approach is to extract the alignment relations from the pretrained model and transfer them to the downstream model, guiding the joint fine-tuning of CLIP's image and text encoders. However, this approach faces a key challenge: as shown in Fig.~\ref{fig:motivation}(a), while the pretrained CLIP learns global image-text alignment, the segmentation task requires fine-grained pixel-text alignment. Specifically, CLIP constructs a coarse semantic space during image-text pretraining~\cite{radford2021clip}, yielding implicit patch-level semantic representations in its intermediate features. However, these representations are inherently noisy~\cite{xie2024sed,cho2024catseg}, resulting in imprecise and ambiguous pixel-text alignment, whereas semantic segmentation demands accurate and deterministic patch-level semantics. Therefore, transferring patch-level alignment information from the pretrained CLIP without denoising is ineffective and can even harm segmentation models. 

Hence, this raises a two-stage challenge: 
\textit{ 1) how to extract refined pixel-level alignment relations from noisy pretrained representations? 2) how to effectively transfer them to guide downstream fine-tuning while preserving modality alignment ?}

In response to these challenges, we propose InfoCLIP, a novel information-theoretic distillation method specifically designed for asymmetric fine-tuning of CLIP.
To address the first challenge, we introduce a Learnable Pixel-Text Alignment Module (LPAM) to extract alignment relations from the pretrained model. By compressing the mutual information between the extracted relations and embeddings of different modalities, the LPAM captures refined modality alignment information. This process denoises CLIP’s local semantic representations, captures key object features, and enhances the overall segmentation perception. To address the second challenge, we maximize the mutual information between the modality alignment relations extracted by the LPAM of the pretrained and fine-tuned models. Leveraging mutual information’s ability to preserve structural information~\cite{zhang2025infosam}, we transfer compact local semantics to protect the modality alignment relations within the fine-tuned model.

Extensive experiments demonstrate that InfoCLIP sets a new state of the art in open-vocabulary semantic segmentation across three benchmarks by fine-tuning CLIP, highlighting its superiority in handling asymmetric vision-language transfer.

Overall, our contribution can be summarized as follows:
\begin{itemize}
    \item We propose InfoCLIP, the first information-theoretic framework for CLIP fine-tuning, introducing a novel distillation strategy tailored to asymmetric transfer of pretrained knowledge, significantly improving performance in open-vocabulary semantic segmentation.
    \item InfoCLIP introduces dual complementary mechanisms for CLIP adaptation: a bottleneck that strategically compresses noisy information in CLIP's local semantic representations, and an adaptive cross-model mutual information maximization scheme that preserves essential modality alignment.
    \item Extensive evaluations on diverse benchmarks show that InfoCLIP consistently surpasses state-of-the-art open-vocabulary semantic segmentation methods.
\end{itemize}

\section{Related Work}
\subsection{Open-Vocabulary Semantic Segmentation}
Existing open-vocabulary semantic segmentation methods leverage the pretrained CLIP~\cite{radford2021clip} to ensure generalization capability. 
However, fine-tuning CLIP on limited training classes impairs its ability to generalize to unseen categories~\cite{zhou2022maskclip}.
To mitigate this issue, mask-based methods~\cite{ghiasi2022openseg,xu2022zsseg,xu2023odise} freeze the pretrained vision-language model and introduce an auxiliary mask generator~\cite{cheng2022masked} for segmentation, thereby preserving open-vocabulary capability. However, without pixel-wise fine-tuning, the feature distribution in CLIP leads to misalignment between localized visual representations and corresponding text embeddings~\cite{liang2023ovseg}. 
In contrast, pixel-based methods~\cite{xu2023san,yu2023fcclip,cho2024catseg} offer a more promising solution by directly fine-tuning CLIP for pixel-level predictions with a carefully crafted parameter space. 
Although these methods improve segmentation performance, the pretrained vision-language alignment is inevitably degraded due to the paradigm shift from image-text to pixel-text alignment. 
Therefore, we propose an information-theoretic framework to preserve essential modality alignment and enhance generalization.

\subsection{Knowledge Distillation}
Knowledge distillation (KD) is designed to transfer knowledge from a large, well-trained teacher model to a smaller, lightweight student model~\cite{hinton2015distilling}, and is later widely used as an effective tool for transferring knowledge from a source task to a target task~\cite{mansourian2025comprehensive}. For open-vocabulary semantic segmentation, MAFT~\cite{jiao2023maft} introduces a self-distillation loss between the frozen CLIP and the fine-tuned IP-CLIP to preserve CLIP’s transferability and mitigate overfitting. MAFT+~\cite{jiao2024maftplus} further proposes a representation compensation strategy that distills multi-scale features, generated through adaptive pooling, from the frozen CLIP to the fine-tuned CLIP, aiming to maintain zero-shot capability during fine-tuning.
However, these approaches are tailored for mask-based methods that distill knowledge at the image level, without addressing the pixel-text alignment problem, making them unsuitable for pixel-based methods. 

\begin{figure*}[t]
\centering
\includegraphics[width=0.9\textwidth]{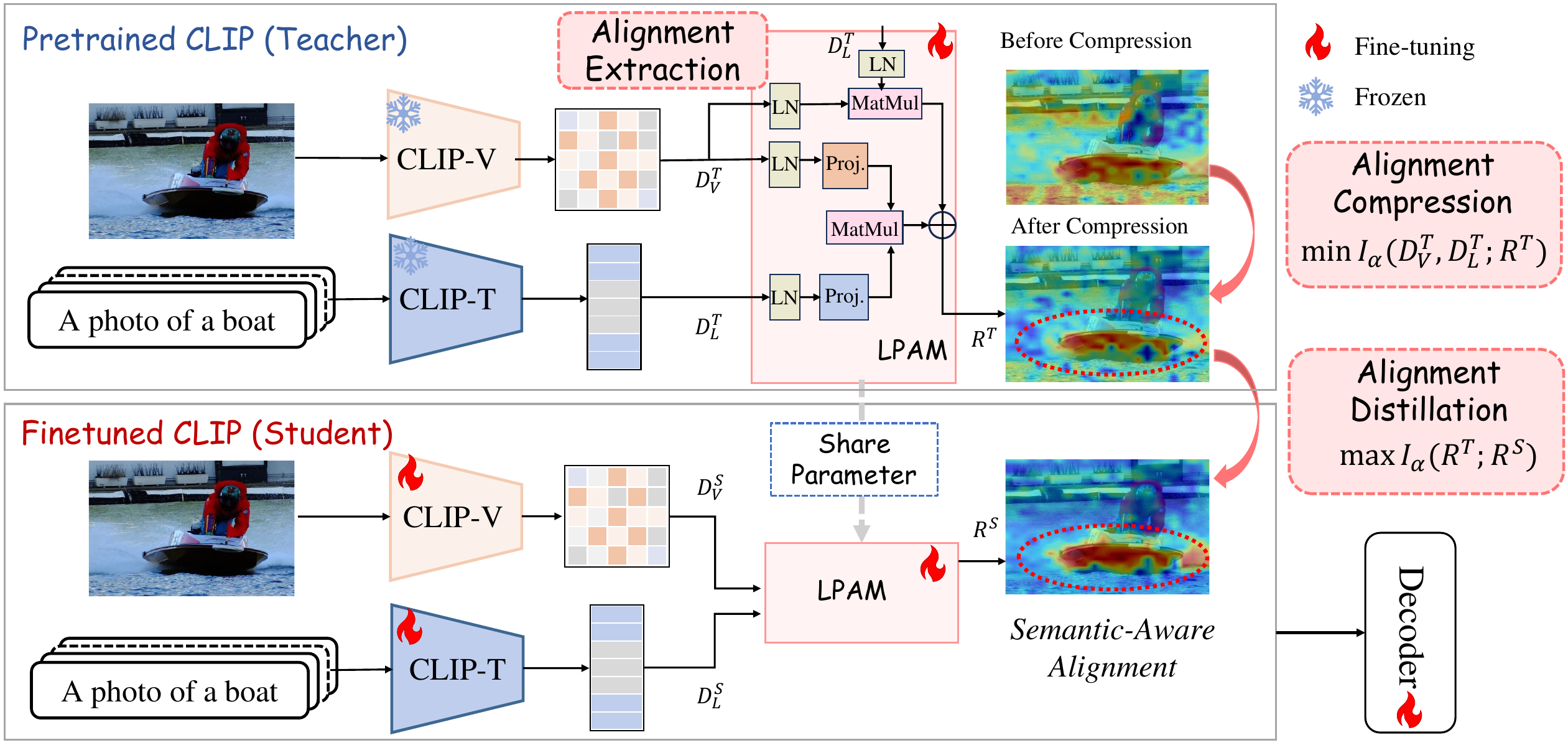}
\caption{\textbf{Overview of InfoCLIP.} To exploit the valuable yet noisy pixel-text alignment from a pretrained foundation model (CLIP) for OVSS, InfoCLIP introduces an information-theoretic framework for asymmetric adaptation, comprising: (1) a Learnable Pixel-Text Alignment Module (LPAM) to extract fine-grained patch-text relations; (2) an information bottleneck loss to suppress noise and retain semantic-aware alignment; and (3) a mutual information transfer loss to preserve modality alignment by bridging pretrained and fine-tuned CLIP representations. The detailed formulation of LPAM is provided in the Appendix.}
\label{fig:overview}
\end{figure*}

\section{Preliminaries}
\label{sec:pre}
In information theory, Rényi's $\alpha$-entropy $H_{\alpha}(X)$ generalizes Shannon’s entropy and is defined for a continuous random variable $X$ with PDF $p(x)$ over a finite set $\mathcal{X}$ as:
\begin{equation}
H_{\alpha}(X) = \frac{1}{1 - \alpha} \log \int_{\mathcal{X}} p^{\alpha}(x) \mathrm{d}x,
\end{equation}
where $H_{\alpha}$ reduces to Shannon’s entropy as $\alpha \to 1$. However, its reliance on the underlying data distribution limits its applicability in high-dimensional settings. To address this, matrix-based Rényi’s $\alpha$-entropy enables direct computation from data without density estimation.

\textbf{Definition 1.} Let $\kappa : \mathcal{X} \times \mathcal{X} \mapsto \mathbb{R}$ be a real-valued positive kernel that is also infinitely divisible~\cite{bhatia2006infinitely}. Given $\{\mathbf{x}_i\}_{i=1}^n \subset \mathcal{X}$, each $\mathbf{x}_i$ being a real-valued scalar or vector, and the Gram matrix $K$ obtained from $K_{ij} = \kappa(\mathbf{x}_i, \mathbf{x}_j)$, a matrix-based analogue to Rényi’s $\alpha$-entropy can be defined as:
\begin{equation}
S_\alpha(A) = \frac{1}{1 - \alpha} \log \left( \mathrm{tr}(A^\alpha) \right) 
= \frac{1}{1 - \alpha} \log \left[ \sum_{i=1}^n \lambda_i^\alpha(A) \right],
\label{eq:entropy}
\end{equation}
where $A_{ij} = \frac{1}{n} \frac{K_{ij}}{\sqrt{K_{ii} K_{jj}}}$ is a normalized kernel matrix and $\lambda_i(A)$ denotes the $i$-th eigenvalue of $A$.

The kernel matrix $\mathbf{A}$ is positive semi-definite with $\mathrm{tr}(\mathbf{A}) = 1$, ensuring all eigenvalues $\lambda_i \in [0, 1]$. Under this setting, matrix-based Rényi’s joint entropy, mutual information $I_\alpha(\mathbf{A}; \mathbf{B})$, and their multivariate extensions can be defined~\cite{yu2019multivariate}.

\textbf{Definition 2.} Let $\kappa_1 : \mathcal{X}^1 \times \mathcal{X}^1 \mapsto \mathbb{R}, \dots, \kappa_L : \mathcal{X}^L \times \mathcal{X}^L \mapsto \mathbb{R}$ be positive, infinitely divisible kernels, and let $\{ \mathbf{x}_i^1, \dots, \mathbf{x}_i^L \}_{i=1}^n \subset \mathcal{X}^1 \times \dots \times \mathcal{X}^L$ be a collection of $n$ samples. A matrix-based analogue to Rényi's $\alpha$-order joint entropy among $L$ variables can be defined as:
\begin{equation}
S_\alpha(A_1, \dots, A_L) = S_\alpha \left( \frac{A_1 \circ \dots \circ A_L}{\mathrm{tr}(A_1 \circ \dots \circ A_L)} \right),
\end{equation}
where $A_1, \dots, A_L$ are normalized kernel matrices, and $\circ$ denotes the Hadamard product. Within the settings, the mutual information between $A_1$ and $A_2$, denoted as $I_\alpha(A_1 ; A_2)$, is given by:
\begin{equation}
\label{eq_mi}
I_\alpha(A_1 ; A_2) = S_\alpha(A_1) + S_\alpha(A_2) - S_\alpha(A_1, A_2).
\end{equation}
This formulation avoids the need for high-dimensional probability density estimation required by Shannon entropy, providing a more accurate and computationally efficient alternative~\cite{yu2019multivariate}.

\section{Methodology}
\subsection{Overview of InfoCLIP}

Given an image $I$ and a set of candidate class categories $\mathcal{C} = \{T^{(n)}\}_{n=1}^{N_C}$, where $T^{(n)}$ denotes the textual description of the $n$-th category and $N_C$ is the number of classes, open-vocabulary semantic segmentation aims to assign a class label to each pixel in image $I$~\cite{cho2024catseg}. 
Unlike classical semantic segmentation tasks~\cite{he2019adaptive,zhou2022rethinking} with a fixed label space, open-vocabulary segmentation poses an additional challenge: the category set $\mathcal{C}$ varies across different images and is defined by arbitrary text descriptions.

Fig.~\ref{fig:overview} illustrates the proposed InfoCLIP, which contains three synergistic components: (1) a Learnable Pixel-Text Alignment Module (LPAM) that extracts fine-grained alignment relations between visual patches and text embeddings from the pretrained CLIP; (2) an information bottleneck loss that compresses the extracted alignment relations to suppress noisy semantics and preserve key semantic-aware information; and (3) a mutual information transfer loss that bridges the pretrained and fine-tuned CLIP by maximizing the mutual information between their alignment representations, ensuring the preservation of modality alignment during asymmetric fine-tuning. 
This information-theoretic design enables InfoCLIP to retain the structured alignment knowledge of pretrained CLIP while enhancing its local semantic precision for open-vocabulary segmentation.

\subsection{Learnable Pixel-Text Alignment Module (LPAM)}
\label{sec:LPAM}
While the CLIP model is pretrained to capture global image-text alignment, its intermediate visual features implicitly encode rich patch-level semantics that are potentially valuable for pixel-wise prediction~\cite{radford2021clip}. 
Motivated by this observation, we design a Learnable Pixel-Text Alignment Module (LPAM) to explicitly extract fine-grained alignment relations between image regions and textual concepts. 

Specifically, given an image $I$ and a set of classes $\mathcal{C}$, we extract dense image embeddings $D_V = \Phi_V(I) \in \mathbb{R}^{(H \times W) \times d}$ and text embeddings $D_L = \Phi_L(T) \in \mathbb{R}^{N_C \times d}$, where $\Phi_V(\cdot)$ and $\Phi_L(\cdot)$ are the image and text encoders of CLIP, respectively. Following prior work~\cite{zhou2022extract,cho2024catseg}, we modify the last attention layer of the image encoder to remove the pooling operation. Based on this, the semantic alignment map $R \in \mathbb{R}^{(H \times W) \times N_C}$ is computed by the proposed LPAM as:
\begin{equation}
R = f_{LPAM}(D_V, D_L),
\end{equation}
where $f_{LPAM}: \mathbb{R}^{(H \times W) \times d} \times \mathbb{R}^{N_C \times d} \rightarrow \mathbb{R}^{(H \times W) \times N_C}$ is a learnable function that computes fine-grained similarities between image patches and class embeddings. As shown in Fig.~\ref{fig:overview}, LPAM aligns each visual token with all class embeddings via a learned attention mechanism, producing a dense alignment map for pixel-level semantic prediction.

Notably, this module is applied to both the pretrained CLIP (teacher) and the fine-tuned CLIP (student), denoted as $f_{LPAM}^T$ and $f_{LPAM}^S$, respectively, with shared parameters. Consequently, their resulting pixel-text alignments are represented as $R^T$ and $R^S$.

\subsection{Semantic Compression via Information Bottleneck}
\label{sec:compress}
While LPAM extracts fine-grained pixel-text alignment maps from CLIP’s intermediate features, these representations originate from a global image-text alignment objective and are not explicitly optimized for dense prediction. As a result, they often contain noisy or entangled semantics that lack the clarity and determinism required for accurate pixel-level segmentation~\cite{xie2024sed,cho2024catseg}.

To address this issue, we propose an information bottleneck mechanism that compresses the alignment map to extract essential semantic-aware semantics while suppressing irrelevant or noisy signals.
Given the dense image embeddings $D_V^T$ and text embeddings $D_L^T$ from the pretrained CLIP, and their corresponding pixel-level alignment map $R^T$ produced by LPAM, our first goal is to regulate the information flow by minimizing the mutual information $\mathbf{I}_\alpha(D_V^{T}, D_L^{T}; \, R^{T})$. Here, $\mathbf{I}_\alpha(\cdot\,;\cdot)$ denotes the $\alpha$-Rényi's mutual information, which measures the amount of information shared between the input embeddings and the alignment map. 

In other words, this can be formulated as a regularization loss to supervise the learning of $R^T$:
\begin{equation}
\begin{aligned}
\mathcal{L}_{c} = &\mathbf{I}_\alpha(D_V^{T}, D_L^{T}; \, R^{T}) \\
       = &\mathbf{S}_\alpha(G^{T}_{V}, G^{T}_{L})+\mathbf{S}_\alpha(G^{T}_R)-\mathbf{S}_\alpha(G^{T}_{V}, G^{T}_{L},G^{T}_R).
\label{eq:loss_c_1}
\end{aligned}
\end{equation}
Notably, the teacher entropy term $\mathbf{S}_\alpha(G^{T}_{V}, G^{T}_{L})$ in this loss can be excluded, as the teacher's weights remain fixed during fine-tuning. The $G^{T}_{V}, G^{T}_{L}, G^{T}_R \in \mathbb{R}^{N \times N}$ are the Gram matrices induced by a batch of normalized features $D_V^{T}$, $D_L^{T}$, and $ R^{T}$ with a polynomial kernel of degree 1~\cite{zhang2025infosam}. For instance, $G^{T}_{V}$ can be defined as follows:
\begin{equation}
\begin{aligned}
 \textcolor{black}{G^{T}_{V} = \frac{\kappa(D_V^{T}, D_V^{T})}{\text{tr}(\kappa(D_V^{T}, D_V^{T}))},}
 \label{eq:gram}
\end{aligned}
\end{equation}
where $\kappa(x, y) = x^\top y$ represents polynomial kernel function and $\text{tr}(\cdot)$ denotes the trace of the matrix. Therefore, according to the Preliminaries and Eq.~\ref{eq:gram}, we can reformulate the compression loss $\mathcal{L}_{c}$ as follows:
\begin{equation}
\begin{aligned}
\label{eq:loss_c_2}
\mathcal{L}_c =& \mathbf{S}_\alpha(G^{T}_R)-\mathbf{S}_\alpha(G^{T}_{V}, G^{T}_{L},G^{T}_R) \\
=& 
\frac{1}{1 - \alpha}  \left[ \log_2 \sum_{i=1}^{n} \lambda_i^\alpha\left(G_{R}^T\right)
-
 \log_2 \sum_{i=1}^{n} \lambda_i^\alpha\left(G_{VLR}^T\right) \right],
\end{aligned}
\end{equation}
where $G_{VLR}^T = G_V^T \circ G_L^T \circ G_R^T$ is derived from the marginal and joint entropy definitions in the Preliminaries, with $\circ$ denoting the Hadamard product.

As computing eigenvalues is expensive~\cite{yu2019multivariate}, we follow prior work~\cite{miles2021information, zhang2025infosam} and set $\alpha = 2$, enabling a Frobenius-norm-based approximation of Rényi’s $\alpha$-entropy: \textcolor{black}{$\|\mathbf{A}\|_F^2 = \mathrm{tr}(\mathbf{A}\mathbf{A}^H) = \sum_{i=1}^n \lambda_i^2(\mathbf{A})$}. Consequently, $\mathcal{L}_c$ can be reformulated as:
\begin{equation}
\begin{aligned}
\label{eq:loss_c}
\mathcal{L}_c = - \log_2 \| G^{T}_{R} \|_F^2 + \log_2 \| G^{T}_{VLR} \|_F^2.
\end{aligned}
\end{equation}
The first term suppresses redundant signals by minimizing the entropy of alignment features, while the second term encourages the alignment map to retain informative joint semantics by maximizing the joint entropy among image, text, and alignment features. Together, they form an information bottleneck that filters noise while preserving key semantics.

\begin{table*}[ht]
\centering
\small
\label{tab:comparison}
\setlength{\tabcolsep}{4pt}
\begin{tabular}{l|c c c|c c c c c c}
\toprule
Model & VLM & Add. Backbone & Training Dataset & A-847 & PC-459 & A-150 & PC-59 & PAS-20 & PAS-20$^b$ \\
\midrule
\midrule
\multicolumn{10}{c}{\textbf{\textit{Without Distillation}}} \\
\midrule
SPNet~\shortcite{xian2019spnet} & - & ResNet-101 & PASCAL VOC & - & - & - & 24.3 & 18.3 & - \\
ZS3Net~\shortcite{bucher2019zs3net} & - & ResNet-101 & PASCAL VOC & - & - & - & 19.4 & 38.3 & - \\
LSeg~\shortcite{li2022lseg} & CLIP ViT-B/32 & ResNet-101 & PASCAL VOC-15 & - & - & - & - & 47.4 & - \\
LSeg+~\shortcite{ghiasi2022openseg} & ALIGN & ResNet-101 & COCO-Stuff & 2.5 & 5.2 & 13.0 & 36.0 & - & 59.0 \\
ZegFormer~\shortcite{ding2022zegformer} & CLIP ViT-B/16 & ResNet-101 & COCO-Stuff-156 & 4.9 & 9.1 & 16.9 & 42.8 & 86.2 & 62.7 \\
ZegFormer~\shortcite{ding2022zegformer} & CLIP ViT-B/16 & ResNet-101 & COCO-Stuff & 5.6 & 10.4 & 18.0 & 45.5 & 89.5 & 65.5 \\
ZSseg~\shortcite{xu2022zsseg} & CLIP ViT-B/16 & ResNet-101 & COCO-Stuff & 7.0 & - & 20.5 & 47.7 & 88.4 & - \\
OpenSeg~\shortcite{ghiasi2022openseg} & ALIGN & ResNet-101 & COCO Panoptic & 4.4 & 7.9 & 17.5 & 40.1 & - & 63.8 \\
OVSeg~\shortcite{liang2023ovseg} & CLIP ViT-B/16 & ResNet-101c & COCO-Stuff & 7.1 & 11.0 & 24.8 & 53.3 & 92.6 & - \\
ZegCLIP~\shortcite{zhou2023zegclip} & CLIP ViT-B/16 & - & COCO-Stuff-156 & - & - & - & 41.2 & 93.6 & - \\
SAN~\shortcite{xu2023san} & CLIP ViT-B/16 & - & COCO-Stuff & 10.1 & 12.6 & 27.5 & 53.8 & 94.0 & - \\
SED~\shortcite{xie2024sed} & CLIP ConvNeXt-B & - & COCO-Stuff & 11.4 & 18.6 & 31.6 & 57.3 & 94.4 & - \\
CAT-Seg~\shortcite{cho2024catseg} & CLIP ViT-B/16 & - & COCO-Stuff & \underline{12.0} & \underline{19.0} & \underline{31.8} & \underline{57.5} & \underline{94.6} & \underline{77.3} \\
\midrule
\multicolumn{10}{c}{\textbf{\textit{With Distillation}}} \\
\midrule
MAFT~\shortcite{jiao2023maft} & CLIP ViT-B/16 & ResNet-101 & COCO-Stuff & 10.1 & 12.8 & 29.1 & 53.5 & 90.0 & - \\
InfoCLIP (ours) & CLIP ViT-B/16 & - & COCO-Stuff & \textbf{12.6} & \textbf{19.5} & \textbf{32.1} & \textbf{58.1} & \textbf{95.5} & \textbf{78.1} \\
\midrule
\midrule
\multicolumn{10}{c}{\textbf{\textit{Without Distillation}}} \\
\midrule
LSeg~\shortcite{li2022lseg} & CLIP ViT-B/32 & ViT-L/16 & PASCAL VOC-15 & - & - & - & - & 52.3 & - \\
OpenSeg~\shortcite{ghiasi2022openseg} & ALIGN & Eff-B7 & COCO Panoptic & 8.1 & 11.5 & 26.4 & 44.8 & - & 70.2 \\
OVSeg~\shortcite{liang2023ovseg} & CLIP ViT-L/14 & Swin-B & COCO-Stuff & 9.0 & 12.4 & 29.6 & 55.7 & 94.5 & - \\
SAN~\shortcite{xu2023san} & CLIP ViT-L/14 & - & COCO-Stuff & 12.4 & 15.7 & 32.1 & 57.7 & 94.6 & - \\
ODISE~\shortcite{xu2023odise} & CLIP ViT-L/14 & Stable Diffusion & COCO-Stuff & 11.1 & 14.5 & 29.9 & 57.3 & - & - \\
SED~\shortcite{xie2024sed} & CLIP ConvNeXt-L & - & COCO-Stuff & 13.9 & 22.6 & 35.2 & 60.6 & 96.1 & - \\
FC-CLIP~\shortcite{yu2023fcclip} & CLIP ConvNeXt-L & - & COCO Panoptic & 14.8 & 18.2 & 34.1 & 58.4 & 95.4 & - \\
CAT-Seg~\shortcite{cho2024catseg} & CLIP ViT-L/14 & - & COCO-Stuff & \underline{16.0} & \underline{23.8} & \underline{37.9} & \underline{63.3} & \underline{97.0} & \underline{82.5} \\
\midrule
\multicolumn{10}{c}{\textbf{\textit{With Distillation}}} \\
\midrule
MAFT~\shortcite{jiao2024maftplus} & CLIP ViT-L/14 & Mask2Former & COCO-Stuff & 12.7 & 16.2 & 33.0 & 59.0 & 92.1 & - \\
MAFT~\shortcite{jiao2024maftplus} & CLIP ConvNeXt-L & Mask2Former & COCO-Stuff & 13.1 & 17.0 & 34.4 & 57.5 & 93.0 & - \\
MAFT+~\shortcite{jiao2024maftplus} & CLIP ConvNeXt-L & Mask2Former & COCO-Stuff & 15.1 & 21.6 & 36.1 & 59.4 & 96.5 & - \\
InfoCLIP (ours) & CLIP ViT-L/14 & - & COCO-Stuff & \textbf{16.6} & \textbf{24.6} & \textbf{38.5} & \textbf{63.5} & \textbf{97.5} & \textbf{83.1} \\
\bottomrule
\end{tabular}
\caption{\textbf{Comparison of open-vocabulary semantic segmentation performance on standard benchmarks.} Bold indicates the best performance, and underlining denotes the second-best.}
\label{tab:main_results}
\end{table*}

\subsection{Alignment Transfer via Mutual Information}
\label{sec:distill}
To improve CLIP's semantic perception in open-vocabulary segmentation, we distill compact semantic-aware pixel-text alignment from the pretrained CLIP into its fine-tuned CLIP. Unlike conventional distillation losses that rely on distribution matching or soft targets, this formulation avoids expensive density estimation and provides a stable, differentiable objective. Specifically, we achieve this distillation by maximizing the mutual information between the extracted semantic alignment representations of the pretrained and fine-tuned models. The distillation loss can be defined as:
\begin{equation}
\begin{aligned}
\label{eq:loss_d_1}
\mathcal{L}_d = & -\mathbf{I}_\alpha(R^{T}; R^{S}) \\
       = &-\mathbf{S}_\alpha(G^{T}_r) - \mathbf{S}_\alpha(G^{S}_R)+\mathbf{S}_\alpha(G^{T}_{R}, G^{S}_{R}),
\end{aligned}
\end{equation}
where $G^{S}_R$ denotes the Gram matrix of $R^{S}$. Following the same derivation as before, the alignment transfer via mutual information can be expressed as:
\begin{equation}
\begin{aligned}
\label{eq:loss_d}
\mathcal{L}_d = \log_2 \| G^{T}_{R} \|_F^2 + \log_2 \| G^{S}_{R} \|_F^2 - \log_2 \| G^{TS}_{R} \|_F^2.
\end{aligned}
\end{equation}
From an optimization perspective, the terms in $\mathcal{L}_d$ serve distinct roles: the first two terms act as regularizers that encourage each model to maintain informative and structured alignment representations, while the third term enforces consistency between the teacher and student via relation-level alignment. This distillation loss $\mathcal{L}_d$ effectively preserves fine-grained, semantically consistent alignment during open-vocabulary segmentation fine-tuning.

\subsection{Overall Training Process}
Following prior work~\cite{cho2024catseg}, we freeze some CLIP layers to reduce computational overhead. For the loss function, we incorporate the task loss (i.e., cross-entropy loss), denoted as $\mathcal{L}_{task}$, for the fine-tuning of CLIP. The overall loss function is derived as:
\begin{equation}
\begin{aligned}
\label{eq_loss_all}
\mathcal{L} = \mathcal{L}_{task} + \lambda_1 \mathcal{L}_{c} + \lambda_2 \mathcal{L}_{d},
\end{aligned}
\end{equation}
where $\lambda_1$ and $\lambda_2$ control the influence of the regularization terms to achieve an effective trade-off between task performance and knowledge preservation. The overall loss $\mathcal{L}$ combines three terms: the task loss $\mathcal{L}_{task}$ for segmentation, the compression loss $\mathcal{L}_{c}$ to preserve key object semantics alignment, and the distillation loss $\mathcal{L}_{d}$ to reinforce the effective semantic alignment learned by the pretrained CLIP into the fine-tuned model, helping to maintain semantic consistency and alleviate overfitting to textual inputs during fine-tuning.

\begin{table*}[ht]
\begin{center}
\begin{small}
\begin{tabular}{l|cc|cccccc}
\toprule
Method & $\mathcal{L}_c$ & $\mathcal{L}_d$ & A-847 & PC-459 & A-150 & PC-59 & PAS-20 & PAS-20$^b$ \\  
\midrule
\midrule
w/o distillation & - & - & 12.0 & 19.0 & 31.8 & 57.5 & 94.6 & 77.3 \\
\midrule
\multicolumn{9}{c}{\textbf{\textit{Distillation Methods}}} \\
\midrule
KL~\shortcite{hinton2015distilling} & - & - & 5.7 & 9.0 & 26.5 & 51.3 & 93.1 & 72.6 \\
MAFT~\shortcite{jiao2023maft} & - & - & 11.5 & 17.8 & 31.8 & 56.4 & 95.2 & 76.7 \\
MAFT+~\shortcite{jiao2024maftplus} & - & - & 11.1 & 17.0 & 30.5 & 55.7 & 94.5 & 76.7 \\
\midrule
\multicolumn{9}{c}{\textbf{\textit{Ours}}} \\
\midrule
\multirow{3}{*}{InfoCLIP}  & \ding{55} & \ding{51} & 11.3 & 18.1 & 30.6 & 57.2 & 94.6 & 77.2 \\
                           & \ding{51} & \ding{55} & 11.8 & 18.5 & 31.5 & 57.2 & 95.1 & 77.1 \\
                           & \ding{51} & \ding{51} & \textbf{12.6} & \textbf{19.5} & \textbf{32.1} & \textbf{58.1} & \textbf{95.5} & \textbf{78.1} \\
\bottomrule
\end{tabular}
\end{small}
\end{center}
\caption{\textbf{Ablation study results of the two losses in InfoCLIP and comparison with other distillation methods.} Both base and teacher models are pretrained CLIP ViT-B/16. Existing distillation methods tend to degrade model performance, whereas InfoCLIP effectively transfers beneficial alignment from the pretrained model to the fine-tuned model.}
\label{tab:ablation_loss_and_distillation}
\end{table*}

\begin{figure}[ht]
\centering
\includegraphics[width=0.47\textwidth]{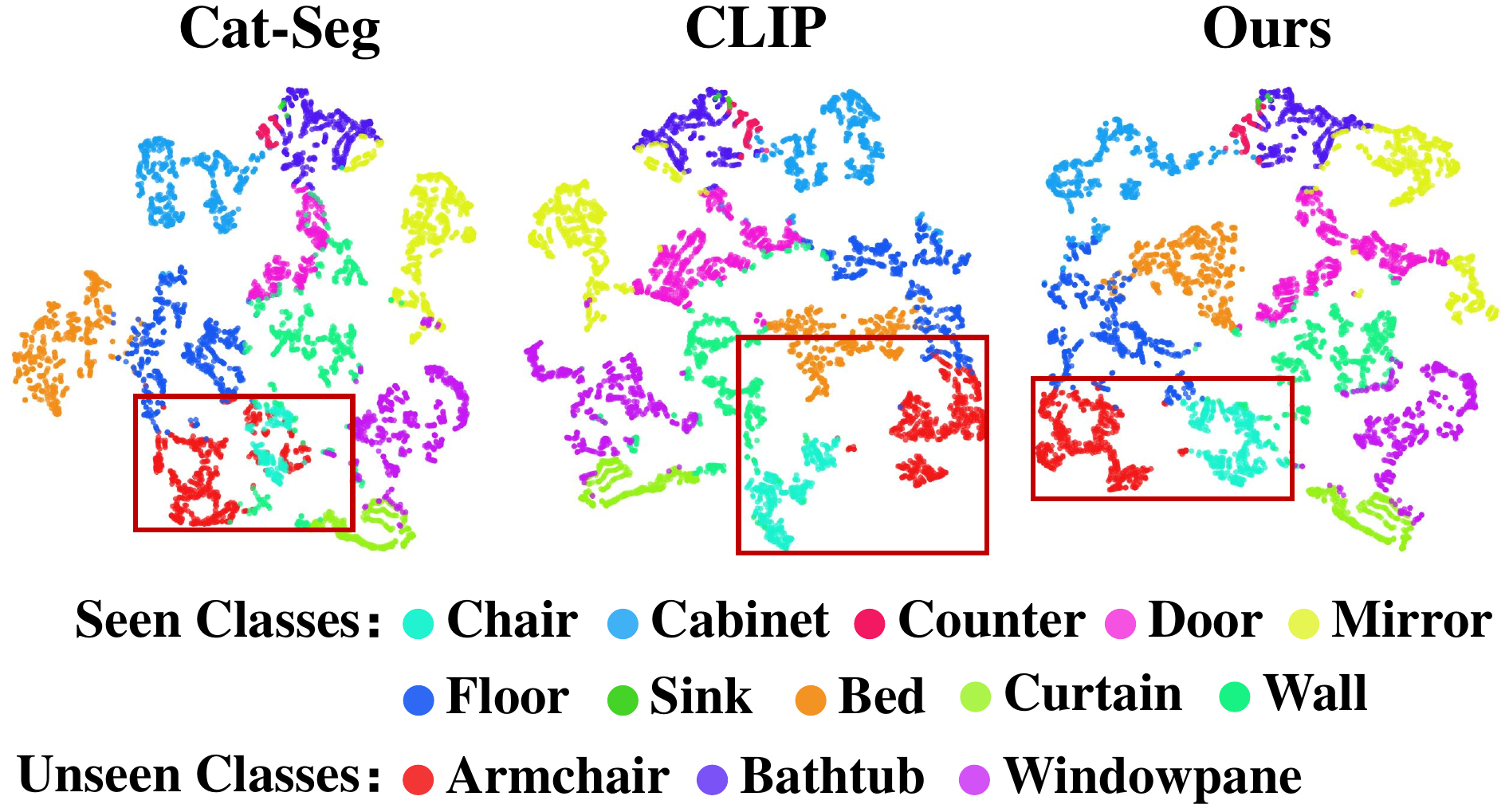}
\caption{\textbf{Effectiveness of alignment distillation.} We present the t-SNE visualization of CLIP image embeddings. As highlighted in the red boxes, while a state-of-the-art method confuses the features of the seen class \textit{chair} and the unseen class \textit{armchair}, our method differentiates them and alleviates overfitting to seen classes, benefiting from the pretrained knowledge distilled from the teacher CLIP model.}
\label{fig:vis_of_tsne}
\end{figure}

\section{Experiments}
\subsection{Experimental Setup}
\noindent \textbf{Datasets.} 
Following prior work~\cite{cho2024catseg,jiao2024maftplus}, we train our model on COCO-Stuff~\cite{caesar2018cocostuff}, which comprises 118k densely annotated training images across 171 categories. 
For evaluating open-vocabulary semantic segmentation performance, we test our model on ADE20K~\cite{zhou2019ade20k}, PASCAL VOC~\cite{everingham2010pascalvoc}, and PASCAL-Context~\cite{mottaghi2014pascalcontext}. 
Specifically, 
ADE20K contains 20k training and 2k validation images annotated with two category sets: A-150, consisting of 150 frequent classes, and A-847, covering 847 total classes. 
PASCAL-Context provides 5k training and validation images with dense annotations for 459 categories (PC-459), with a commonly used subset of the 59 most frequent classes (PC-59). 
PASCAL VOC comprises 1.5k training and validation images labeled with 20 foreground object categories (PAS-20) and a background class. We also report results on PAS-20$^b$, where the background is redefined to include classes in PC-59 but not in PAS-20~\cite{cho2024catseg}.

\noindent \textbf{Evaluation metric.} 
To quantitatively evaluate performance, we follow standard practice~\cite{cho2024catseg,jiao2024maftplus} and evaluate semantic segmentation results using mean Intersection over Union (mIoU).

\noindent \textbf{Implementation Details.} 
In line with previous studies~\cite{cho2024catseg}, we evaluate our method using two Transformer-based CLIP variants, ViT-B/16 and ViT-L/14, with the same decoder as CAT-Seg. For training, the AdamW optimizer~\cite{loshchilov2017adamw} is employed with a learning rate of $2 \times 10^{-4}$ for both the decoder~\cite{cho2024catseg} and our distillation module, and $2 \times 10^{-6}$ for the CLIP backbone, along with a weight decay of $10^{-4}$. We set the batch size to 4 and train the model for 80k iterations on a single NVIDIA A800 (80 GB) GPU. The key hyperparameters $\lambda_1$ and $\lambda_2$ are set to 1.

\begin{figure*}[ht]
\centering
\includegraphics[width=0.75\textwidth]{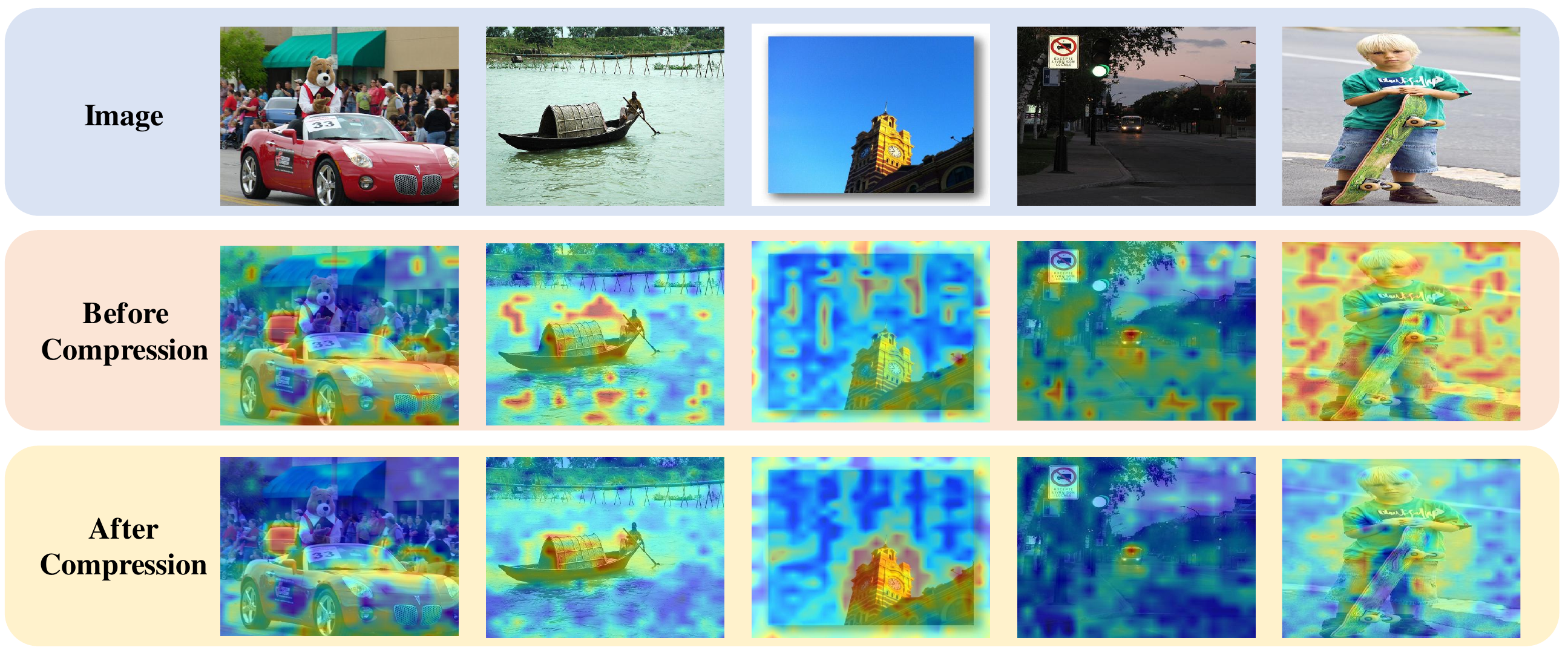}
\caption{\textbf{Effectiveness of semantic alignment extraction and compression.} Semantic compression denoises the pixel-text alignments extracted from the pretrained model, resulting in a sharper focus on the semantic center. From left to right: examples corresponding to the classes \textit{car}, \textit{boat}, \textit{building-other}, \textit{bus}, and \textit{person}.}
\label{fig:vis_relation}
\end{figure*}

\begin{figure}[ht]
\centering
\begin{subfigure}[b]{0.22\textwidth}
    \centering
    \includegraphics[width=\textwidth]{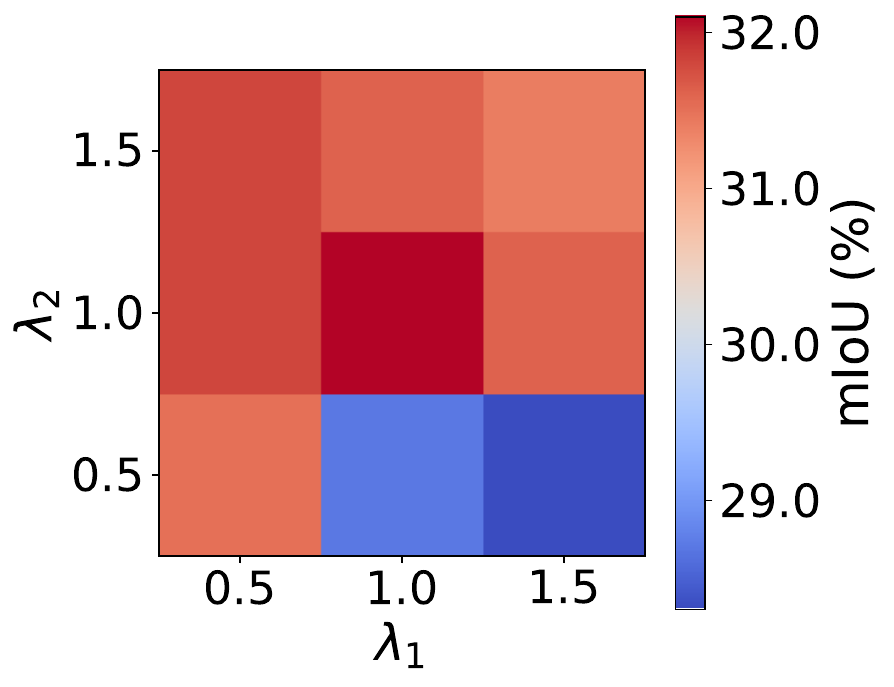}
    \caption{A-150}
    \label{fig:hyper_ablation_a_150}
\end{subfigure}
\hfill
\begin{subfigure}[b]{0.22\textwidth}
    \centering
    \includegraphics[width=\textwidth]{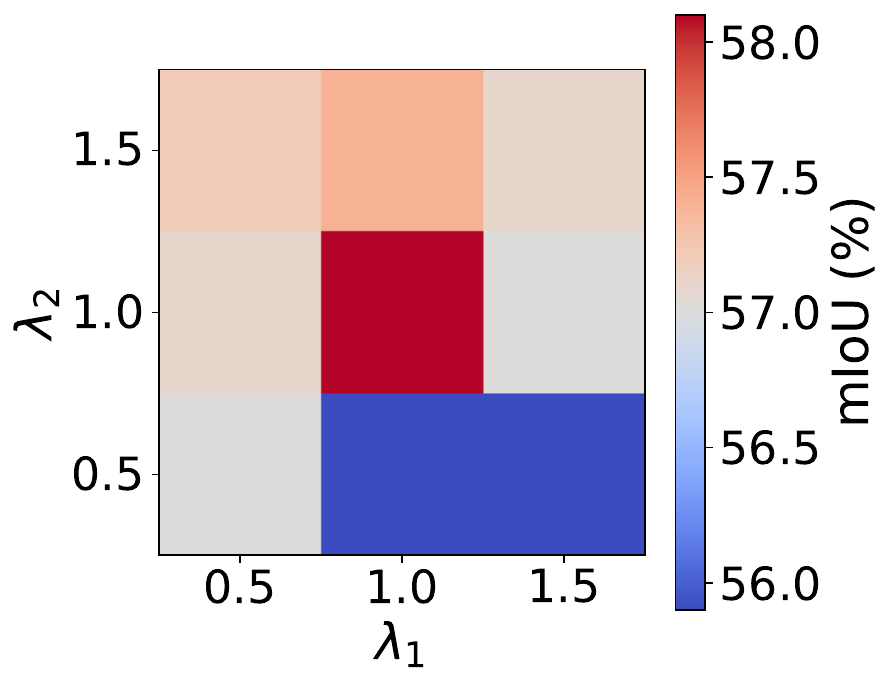}
    \caption{PC-59}
    \label{fig:hyper_ablation_pc_59}
\end{subfigure}
\caption{\textbf{Hyperparameter sensitivity analysis of $\lambda_1$ and $\lambda_2$ balancing $\mathcal{L}_c$ and $\mathcal{L}_d$} on the A-150 and PC-59 datasets.}
\label{fig:hyper_ablation}
\end{figure}

\subsection{Main Results}
\noindent \textbf{Comparison on Standard Benchmarks.} 
Here, we compare our proposed InfoCLIP with several state-of-the-art methods over six test sets across three benchmarks, as summarized in Table~\ref{tab:main_results}. Overall, InfoCLIP achieves the best performance. Among these methods, CAT-Seg~\cite{cho2024catseg} achieves comparable performance to ours. However, it primarily emphasizes local features and employs a complex decoder to reduce redundancy, yet overlooks the degradation of alignment induced by the shift from image-text to pixel-text correspondence. Furthermore, unlike the MAFT series~\cite{jiao2023maft,jiao2024maftplus} that distill visual features, InfoCLIP focuses on pixel-text alignment and achieves superior performance by explicitly mitigating alignment shifts during fine-tuning. Specifically, compared to MAFT, InfoCLIP achieves significant improvements of 8.4\% on the PC-459 dataset and 4.5\% on the PC-59 dataset when using ViT-L/14 as the backbone. These results demonstrate the effectiveness of our method in preserving generalization while learning pixel-level alignment semantic segmentation knowledge.
Furthermore, we present an efficiency analysis in the Appendix.

\noindent \textbf{Qualitative Analysis.} To further validate the effectiveness of InfoCLIP, we present the t-SNE~\cite{maaten2008visualizing} visualization of the dense image embeddings from CLIP on the A-150~\cite{zhou2019ade20k} dataset. Following prior work~\cite{cho2024catseg}, we color the embeddings according to the predicted text classes. 
As shown in Fig.~\ref{fig:vis_of_tsne}, outlined in red boxes, without preserving the pretrained alignment knowledge, the state-of-the-art method tends to overfit to seen classes during fine-tuning. Specifically, as the seen class \textit{chair} is reinforced during fine-tuning, Cat-Seg often misclassifies similar objects, such as the unseen class \textit{armchair}, as \textit{chair}, resulting in entangled features between the two classes. Fortunately, the pretrained CLIP model establishes a distinctive feature space through large-scale vision-language pretraining. Our proposed InfoCLIP effectively transfers this alignment knowledge to the fine-tuned model, successfully disentangling the feature spaces of \textit{chair} and \textit{armchair}. Furthermore, we visualize the prediction results and provide detailed illustrations in the Appendix.

\subsection{Ablation Studies}

\noindent \textbf{Ablation of Main Components.}
Here, we conduct an ablation study to demonstrate the benefits of each component of our proposed InfoCLIP: semantic alignment compression loss $\mathcal{L}_c$ and alignment distillation loss $\mathcal{L}_d$, as shown in Table~\ref{tab:ablation_loss_and_distillation}. We use the ViT-B/16 version of CLIP as the backbone. Additionally, we implement several representative distillation methods, including Kullback-Leibler (KL) divergence-based distillation and the distillation methods from the MAFT series~\cite{jiao2023maft,jiao2024maftplus}, as shown in rows 2 to 4. Notably, existing distillation-based methods even underperform compared to their non-distilled counterparts. In contrast, InfoCLIP extracts generalized semantic-level pixel-text alignment from the pretrained CLIP and suppresses noise through information compression, enabling effective knowledge transfer to the fine-tuned CLIP and leading to improved performance. To ablate the proposed loss functions, row 5 applies a mutual information-based distillation loss on cost volumes~\cite{cho2024catseg}, while row 6 replaces the mutual information loss $\mathcal{L}_d$ with a KL divergence loss. The results show that combining both losses yields additional improvements, with a 1.4\% gain on the PC-459 dataset. This highlights the necessity of extracting clean and informative signals from the pretrained CLIP to effectively suppress the impact of noise.

\noindent \textbf{Discussion of Semantic Alignment Extraction and Compression.}
Furthermore, we visualize and evaluate the effectiveness of the Learnable Pixel-Text Alignment Module (LPAM) and the subsequent information bottleneck compression imposed on its outputs via the loss function $\mathcal{L}_c$. As shown in Fig.~\ref{fig:vis_relation}, the second row, marked with an orange background, presents the heatmap of the cost volume computed via cosine similarity~\cite{cho2024catseg}, whereas the third row illustrates the outputs of the proposed InfoCLIP’s learnable alignment module with compression. The results indicate that our approach more effectively captures prompt-guided object semantic information, yielding clearer and more accurate pixel-level alignment.

\noindent \textbf{Analysis of Hyperparameters $\lambda_1$ and $\lambda_2$.} In Fig.~\ref{fig:hyper_ablation}, we present a key hyperparameter sensitivity analysis of \(\lambda_1\) and \(\lambda_2\), which balance the compression loss \(\mathcal{L}_c\) and the distillation loss \(\mathcal{L}_d\) across multiple benchmarks. Each subfigure displays the mIoU heatmap under varying hyperparameter settings, illustrating the impact on the overall loss. Based on the heatmap, we select \(\lambda_1 = 1\) and \(\lambda_2 = 1\) as the best-performing configuration. Please refer to the Appendix for further results and detailed analysis.

\section{Conclusion}
We present InfoCLIP, an information-theoretic framework featuring two complementary modules for extracting and transferring semantic alignment, tailored for the asymmetric fine-tuning of CLIP. InfoCLIP extracts denoised, task-relevant pixel-text alignment from pretrained CLIP representations and transfers this refined knowledge to downstream segmentation tasks. By compressing and preserving mutual information, InfoCLIP bridges the gap between CLIP’s coarse-grained pretraining objectives and the fine-grained requirements of semantic segmentation, leading to more accurate and robust alignment. Extensive experiments demonstrate that InfoCLIP consistently outperforms prior methods across multiple benchmarks, establishing a new state of the art and highlighting the promise of information-driven alignment transfer in vision-language models.

\section*{Acknowledgements}
This work was supported by the National Natural Science Foundation of China under Grants 62192781, 62172326, 62137002, and 62302384, the Key Research and Development Project in Shaanxi Province No. 2023GXLH-024, and the Project of China Knowledge Centre for Engineering Science and Technology.

\bibliography{aaai2026}

\newpage
\appendix
In the following appendix, we:
\begin{itemize}
  \item present more implementation details of the proposed InfoCLIP;
  \item report additional experimental results on training efficiency and ablation studies;
  \item provide qualitative visualizations;
  \item offer further discussion of the InfoCLIP framework.
\end{itemize}

\section{Implementation Details}

\subsection{Overall Architecture}
We implement our InfoCLIP based on prior work, CAT-Seg~\cite{cho2024catseg}.
Specifically, CAT-Seg comprises a vision encoder, a text encoder, and a cost aggregation module (decoder) that constructs aggregated cost maps to guide open-vocabulary segmentation. 
Built upon this foundation, InfoCLIP adopts a frozen CLIP~\cite{radford2021clip} as a teacher model, distilling compact local semantic relations to the fine-tuned model for more stable cross-modal alignment. 

\begin{figure}[hb]
\centering
\includegraphics[width=0.4\textwidth]{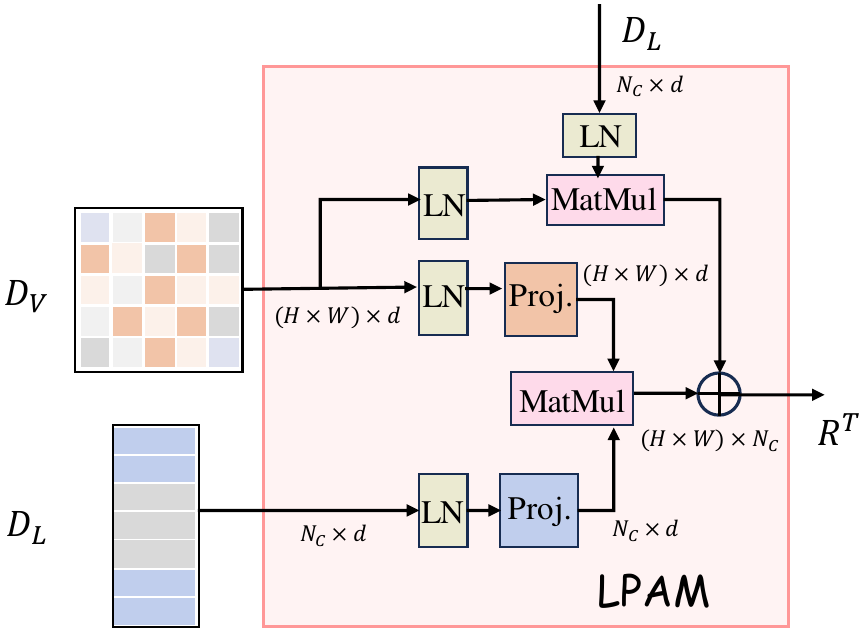}
\caption{\textbf{Architecture of the attention-based Learnable Pixel-Text Alignment Module (LPAM).} LPAM is designed to capture fine-grained alignments between image regions and textual concepts, enabling effective cross-modal interaction within CLIP.}
\label{fig_appendix:LPAM}
\end{figure}

\begin{figure*}[ht]
\centering
\begin{subfigure}[b]{0.3\textwidth}
    \centering
    \includegraphics[width=\textwidth]{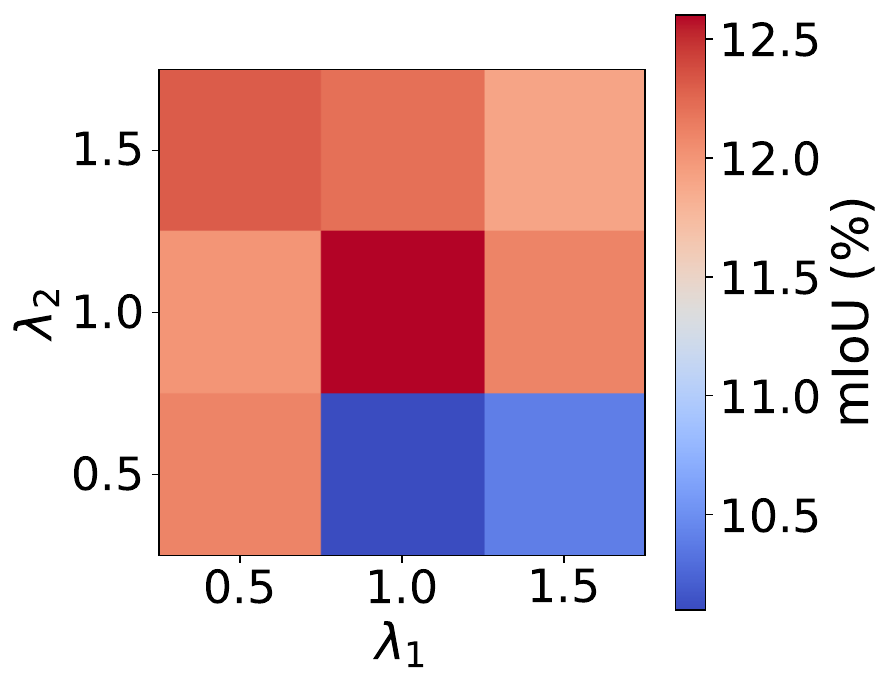}
    \caption{A-847}
    \label{fig_appendix:hyper_ablation_a_847}
\end{subfigure}
\hfill
\begin{subfigure}[b]{0.3\textwidth}
    \centering
    \includegraphics[width=\textwidth]{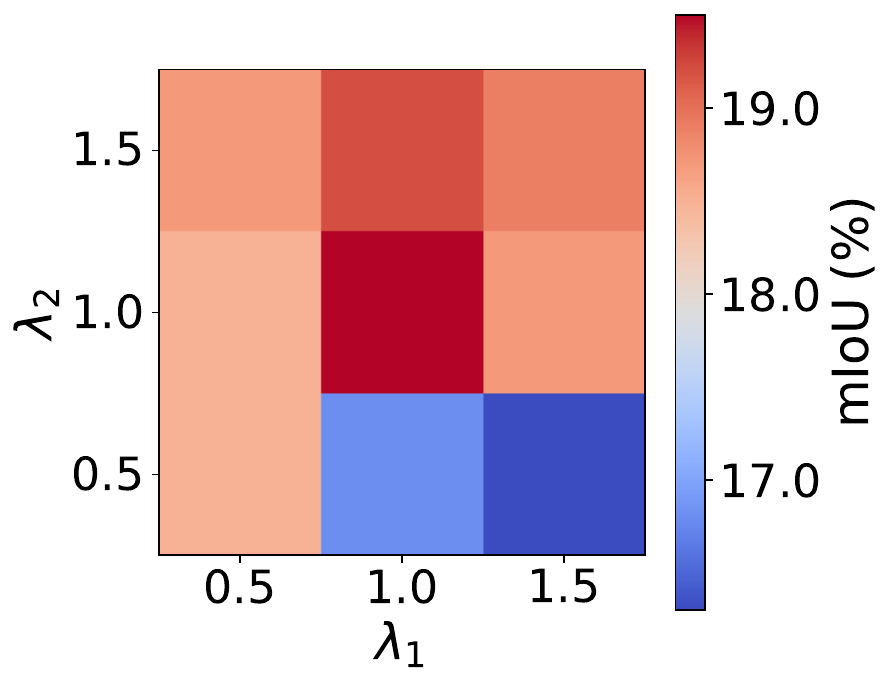}
    \caption{PC-459}
    \label{fig_appendix:hyper_ablation_pc_459}
\end{subfigure}
\hfill
\begin{subfigure}[b]{0.3\textwidth}
    \centering
    \includegraphics[width=\textwidth]{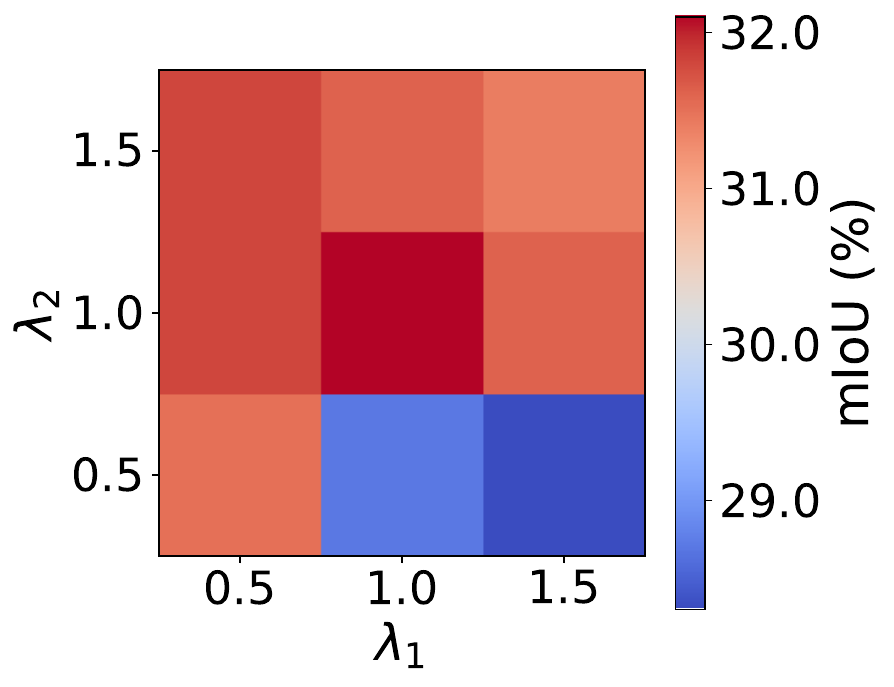}
    \caption{A-150}
    \label{fig_appendix:hyper_ablation_a_150}
\end{subfigure}

\begin{subfigure}[b]{0.3\textwidth}
    \centering
    \includegraphics[width=\textwidth]{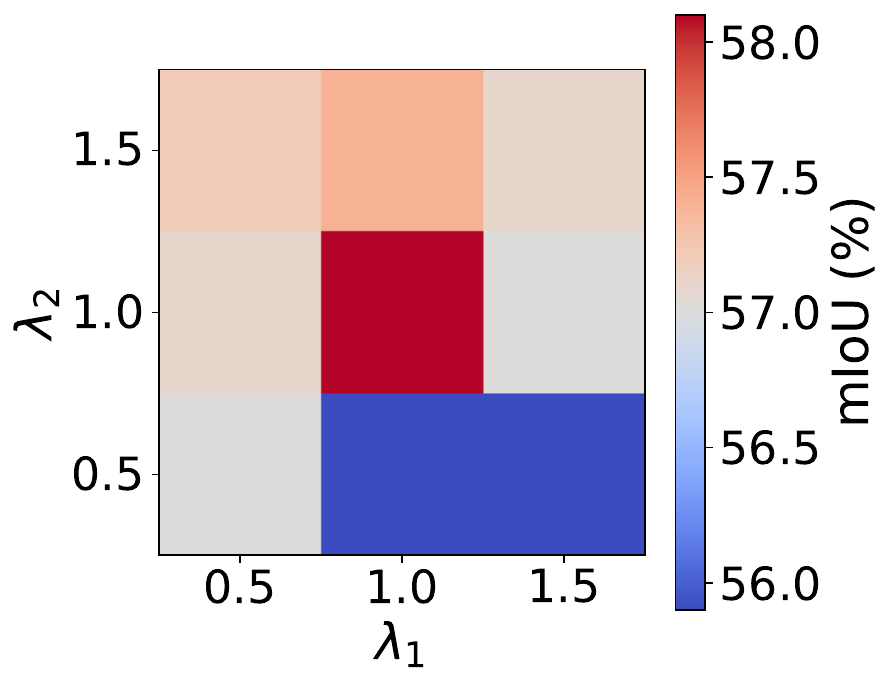}
    \caption{PC-59}
    \label{fig_appendix:hyper_ablation_pc_59}
\end{subfigure}
\hfill
\begin{subfigure}[b]{0.3\textwidth}
    \centering
    \includegraphics[width=\textwidth]{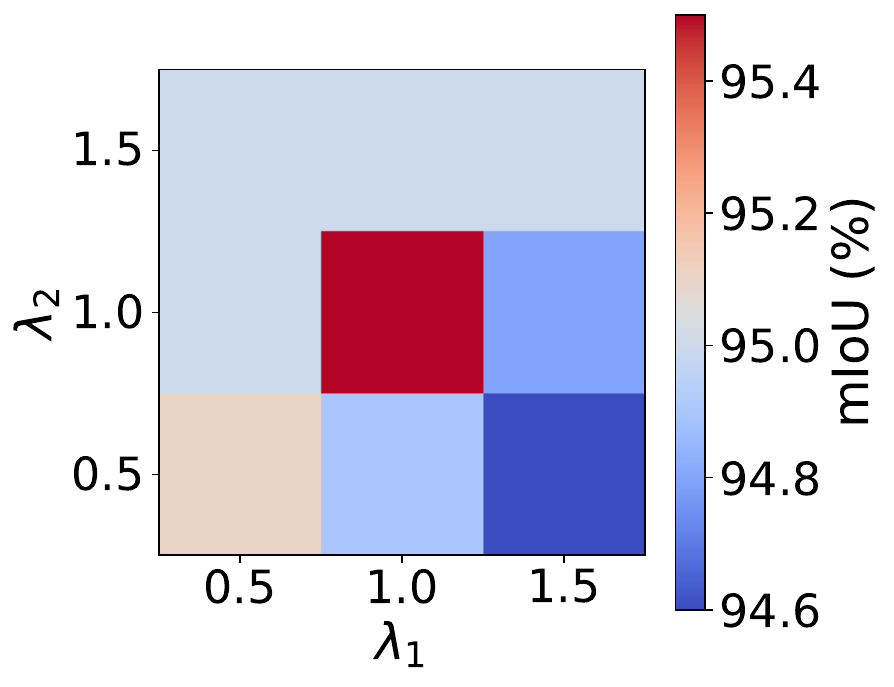}
    \caption{PAS-20}
    \label{fig_appendix:hyper_ablation_pas_20}
\end{subfigure}
\hfill
\begin{subfigure}[b]{0.3\textwidth}
    \centering
    \includegraphics[width=\textwidth]{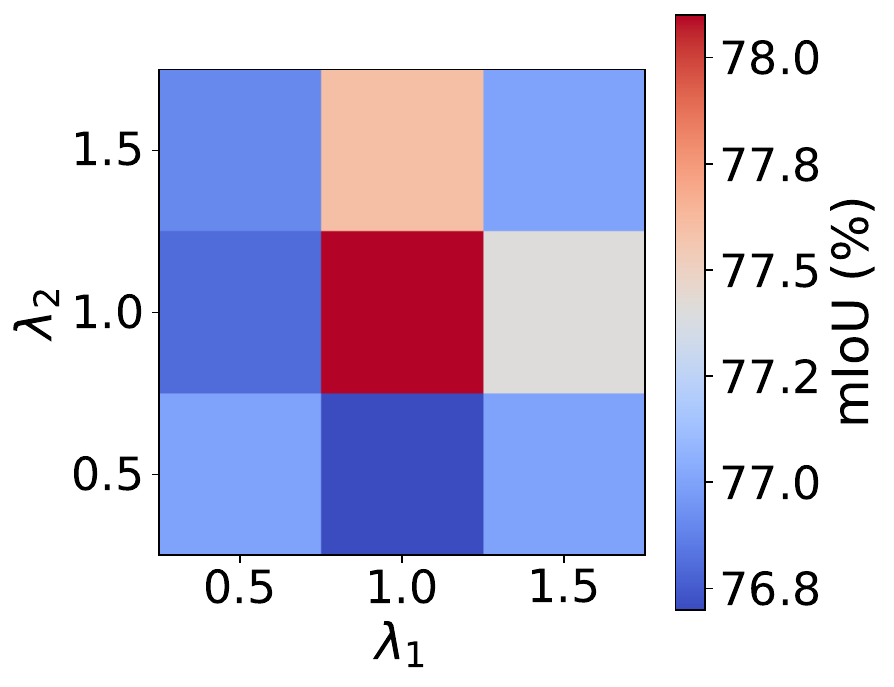}
    \caption{PAS-20$^b$}
    \label{fig_appendix:hyper_ablation_pas_20b}
\end{subfigure}

\caption{\textbf{Hyperparameter sensitivity analysis of \(\lambda_1\) and \(\lambda_2\) balancing \(\mathcal{L}_c\) and \(\mathcal{L}_d\)} across six test sets. The configuration \(\lambda_1 = 1\) and \(\lambda_2 = 1\) consistently yields the best performance.}

\label{fig_appendix:hyper_ablation}
\end{figure*}

\begin{table*}[ht]
\begin{center}
\begin{small}
\begin{tabular}{l|ccc|ccc}
\toprule
Model & Params & \makecell[c]{Teacher\\Params} & \makecell[c]{Training\\Params} & \makecell[c]{CUDA Time\\(Forward, s)} & \makecell[c]{CUDA Time\\(Backward, s)} & \makecell[c]{Peak Memory\\Usage (MB)} \\
\midrule
CAT-Seg & 154.29M & - & 25.11M & 0.37 $\pm$ 0.02 & 0.53 $\pm$ 0.02 & 42111 \\
\midrule
InfoCLIP & 154.82M & 149.62M & 25.64M & 0.45 $\pm$ 0.03 & 0.56 $\pm$ 0.03 & 42714 \\
\bottomrule
\end{tabular}
\end{small}
\end{center}
\caption{\textbf{Analysis of computational efficiency.} CLIP ViT-B/16 serves as the backbone in all experiments. CUDA time is reported as the mean and standard deviation for a single forward and backward pass separately. With the teacher model frozen and the lightweight LPAM containing only 0.53M parameters, InfoCLIP introduces minimal additional computational overhead.}
\label{tab:model_efficiency}
\end{table*}

\begin{table*}[ht]
\begin{center}
\begin{small}
\begin{tabular}{l|cccccc|c}
\toprule
Method & A-847 & PC-459 & A-150 & PC-59 & PAS-20 & PAS-20$^b$ & Comp.~Time (ms) \\  
\midrule
$\alpha=1.01$ & 11.7 & 19.3 & 30.9 & 57.5 & 94.9 & 77.5 & 28.2 $\pm$ 1.5\\
$\alpha=2$    & \textbf{12.6} & \textbf{19.5} & \textbf{32.1} & \textbf{58.1} & \textbf{95.5} & \textbf{78.1} & \textbf{0.5 $\pm$ 0.1}  \\
$\alpha=3$    & 12.0 & 18.9 & 31.5 & 57.1 & 95.0 & 77.0 & 28.1 $\pm$ 1.1 \\
\bottomrule
\end{tabular}
\end{small}
\end{center}
\caption{\textbf{Ablation analysis of the impact of varying $\alpha$ values in the matrix-based Rényi entropy.} When $\alpha = 2$, the computation cost is significantly reduced by computing the matrix-based Rényi entropy using the Frobenius norm, while also achieving the best performance.}

\label{tab:ablation_alpha}
\end{table*}

\subsection{Detailed Formulation of LPAM}
To capture fine-grained alignment between image regions and textual concepts, we propose the Learnable Pixel-Text Alignment Module (LPAM). As shown in Fig.~\ref{fig_appendix:LPAM}, given dense image embeddings $D_V \in \mathbb{R}^{(H \times W) \times d}$ from the image encoder and text embeddings $D_L \in \mathbb{R}^{N_C \times d}$ from the text encoder, LPAM computes a semantic alignment map $R \in \mathbb{R}^{(H \times W) \times N_C}$, where each entry reflects the relevance between an image patch and a category token.
We first normalize the input features using Layer Normalization:
\begin{equation}
\begin{aligned}
D_V^{\text{norm}} &= \text{LayerNorm}(D_V), \\
D_L^{\text{norm}} &= \text{LayerNorm}(D_L).
\end{aligned}
\end{equation}
The normalized features are then projected into query and key spaces:
\begin{equation}
\begin{aligned}
Q &= W_Q \cdot D_V^{\text{norm}}, \\
K &= W_K \cdot D_L^{\text{norm}},
\end{aligned}
\end{equation}
where $W_Q, W_K \in \mathbb{R}^{d \times d}$ are learnable projection matrices.
To enhance alignment robustness, the semantic alignment map $R$ is computed by integrating scaled dot-product attention with a residual similarity term derived from the direct dot product of normalized features:
\begin{equation}
R = \frac{Q K^\top}{\sqrt{d}} + D_V^{\text{norm}} D_L^{\text{norm}^\top}.
\end{equation}
This hybrid formulation yields a robust and effective pixel-text alignment.

\section{Additional Experiment Results}

\subsection{Efficiency Analysis} 
To show the additional overhead of our proposed InfoCLIP, we dive into the training process and provide an analysis of computational efficiency in Table~\ref{tab:model_efficiency}.
It is worth noting that although we introduce a CLIP teacher, it remains frozen during training, so no gradients pass through the teacher model, resulting in only 0.03s of additional CUDA time for a single backward pass.

\subsection{Sensitivity Analysis of \(\lambda_1\) and \(\lambda_2\)}
In Fig.~\ref{fig_appendix:hyper_ablation}, we conduct a comprehensive sensitivity analysis of the key hyperparameters \(\lambda_1\) and \(\lambda_2\) across multiple benchmarks~\cite{zhou2019ade20k, everingham2010pascalvoc, mottaghi2014pascalcontext}. All models are based on CLIP ViT-B/16. Each subfigure presents the mIoU heatmap under varying hyperparameter settings, illustrating their impact on the overall loss. For best clarity, viewing the figure in color is recommended. Consistently across all test sets, the configuration \(\lambda_1 = 1\) and \(\lambda_2 = 1\) achieves the best performance.

\subsection{Analysis of the $\alpha$ in Matrix-based Rényi's Entropy}
In our implementation, the parameter $\alpha$ in the matrix-based Rényi's entropy is set to 2. This choice is primarily motivated by two factors: computational efficiency and theoretical advantages.

Firstly, setting $\alpha$ to 2 enables the computation of the matrix-based Rényi entropy using Frobenius norm operations, thereby eliminating the need for eigenvalue decomposition. This optimization reduces the time complexity from $\mathcal{O}(n^3)$ to $\mathcal{O}(n^2)$ (where $n$ denotes the number of samples)~\cite{dong2023optimal}, substantially lowering computational cost while preserving theoretical rigor, which is particularly beneficial for high-dimensional data analysis~\cite{yu2019multivariate}. Moreover, the differentiability and strong convexity of the Frobenius norm ensure rapid convergence in gradient-based optimization algorithms~\cite{boyd2004convex}.

Secondly, from a theoretical perspective, if the application requires emphasizing the tails of the distribution (rare events) or multimodal distributions (with multiple peaks), the parameter $\alpha$ should be less than 2 and approach 1 from above. Conversely, if the goal is to highlight the dominant mode (the most probable region), $\alpha$ should be greater than 2 to emphasize central tendencies. Setting $\alpha = 2$ provides a neutral weighting scheme~\cite{yu2019multivariate}. 

To validate our argument, we conduct an analysis evaluating the performance across different $\alpha$ values in Table~\ref{tab:ablation_alpha}. All models are based on CLIP ViT-B/16. The performance is evaluated using mIoU, and the computation time per iteration is reported as the mean and standard deviation over a single training run. Following prior work~\cite{yu2019multivariate}, we set $\alpha = 1.01$ to asymptotically approximate Shannon entropy. The results demonstrate that $\alpha = 2$ achieves the highest performance across all the benchmarks while reducing computational overhead by an order of magnitude.

\section{Visualization Results}
We present visualization results of mask predictions across various datasets in comparison with the state-of-the-art model, CAT-Seg. These results further highlight the superior performance of the proposed InfoCLIP.

\begin{figure*}[htp]
\centering
\includegraphics[width=0.85\textwidth]{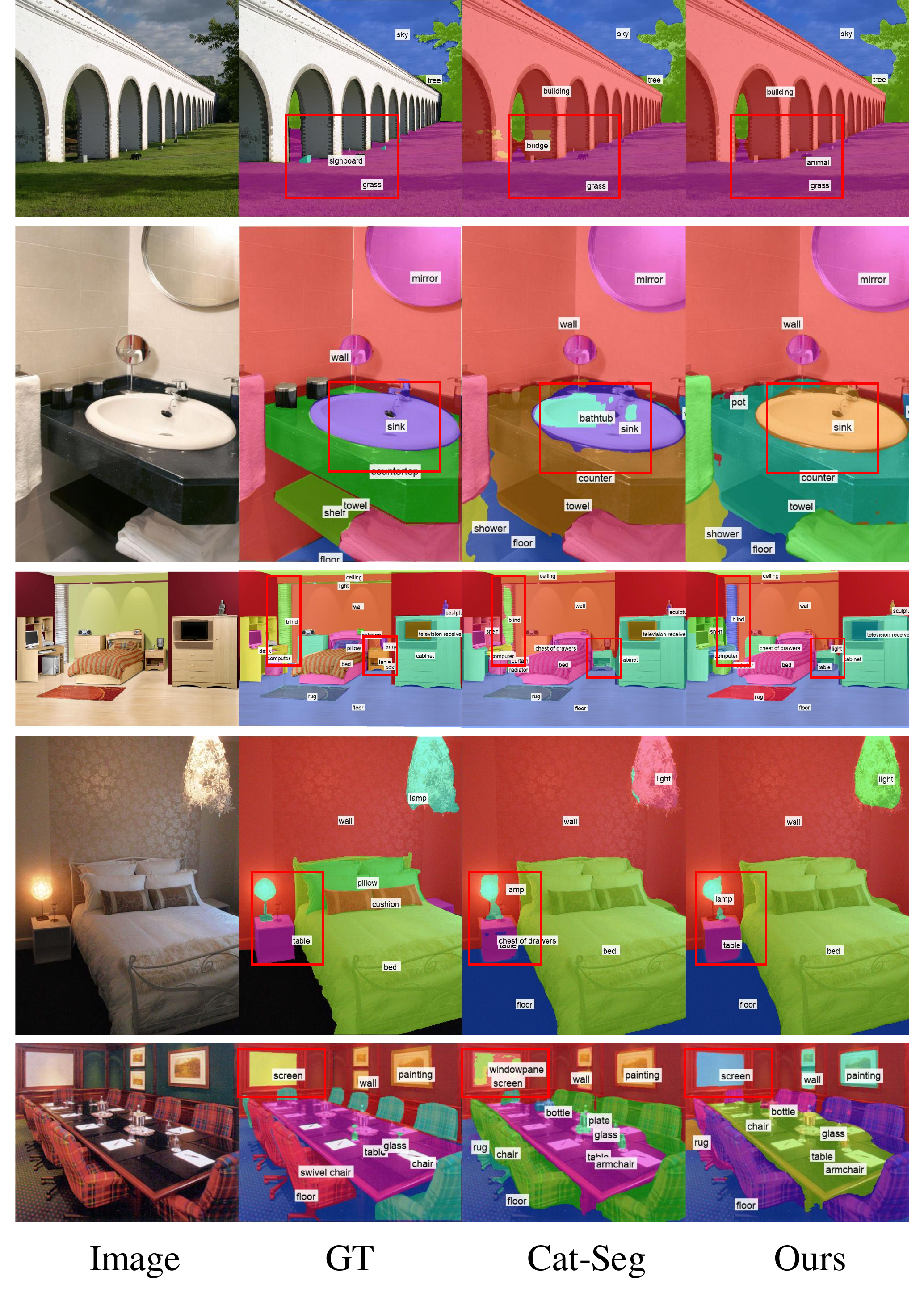}
\caption{Prediction results on ADE20K~\cite{zhou2019ade20k} with 150 categories.}
\label{fig_appendix:mask_A150}
\end{figure*}

\begin{figure*}[htp]
\centering
\includegraphics[width=0.85\textwidth]{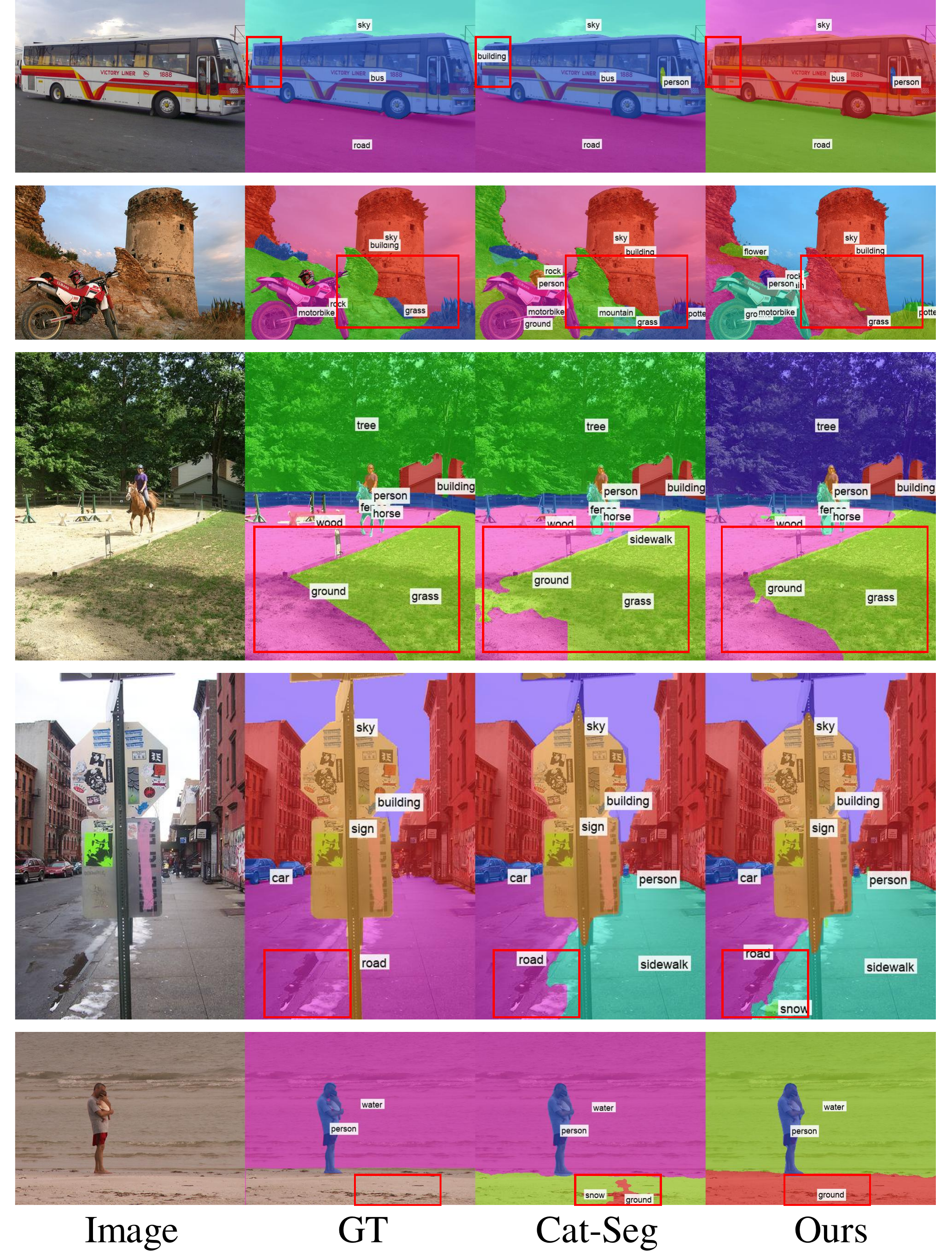}
\caption{Prediction results on PASCAL Context~\cite{mottaghi2014pascalcontext} with 59 categories.}
\label{fig_appendix:mask_PC59}
\end{figure*}

\begin{figure*}[htp]
\centering
\includegraphics[width=0.85\textwidth]{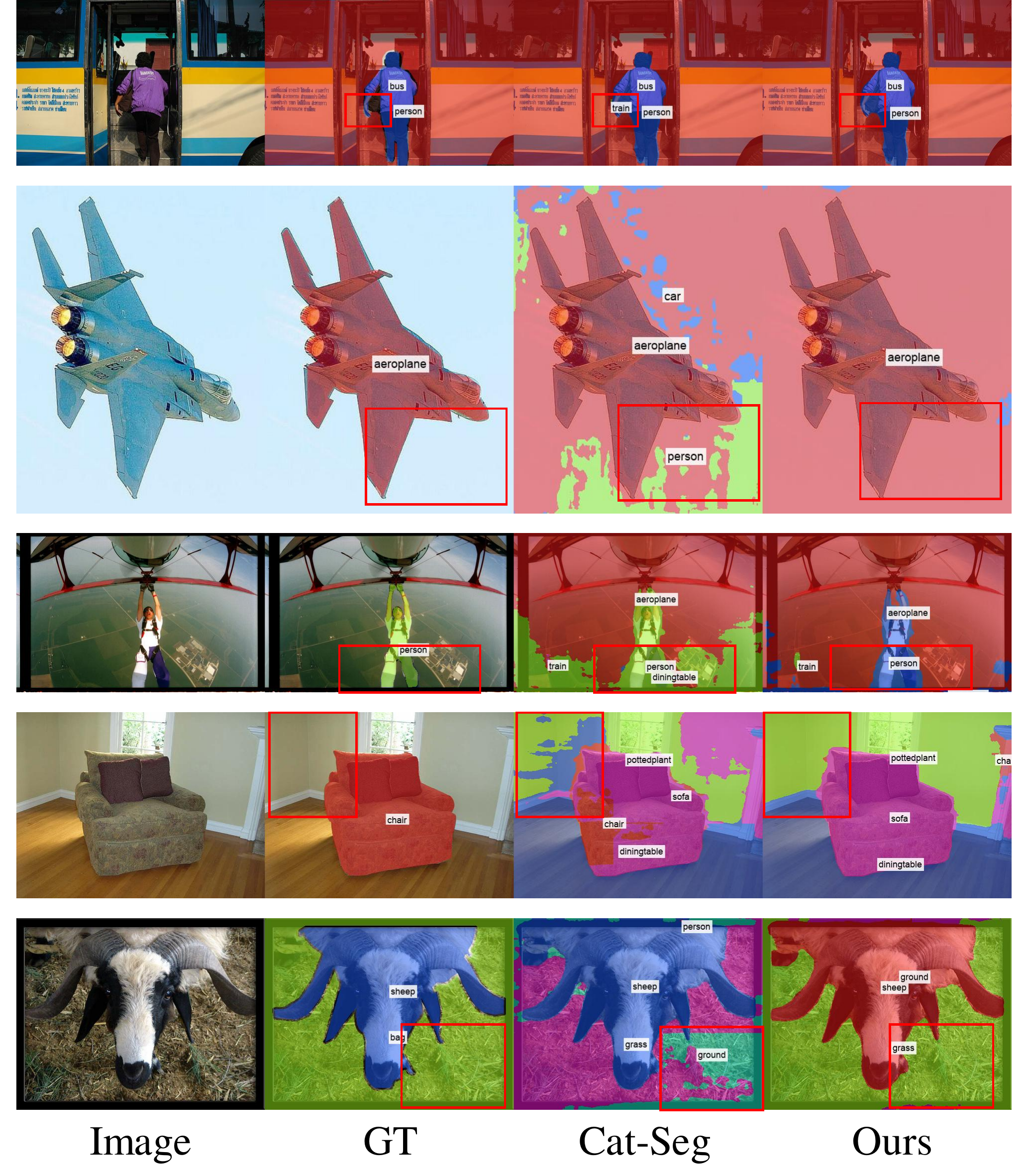}
\caption{Prediction results on PASCAL VOC~\cite{everingham2010pascalvoc} with 20 categories.}
\label{fig_appendix:mask_PAS20}
\end{figure*}

Specifically, we provide qualitative results on A-150~\cite{zhou2019ade20k} in Fig.~\ref{fig_appendix:mask_A150}, PC-59~\cite{mottaghi2014pascalcontext} in Fig.~\ref{fig_appendix:mask_PC59}, and PAS-20~\cite{everingham2010pascalvoc} in Fig.~\ref{fig_appendix:mask_PAS20}. The salient improvements between predictions are indicated by red boxes.

\section{Discussion}

\subsection{Theoretical Tools and Their Role in InfoCLIP} 
InfoCLIP uses information theory to quantify and control cross-modal semantic information in CLIP, keeping task-relevant alignments while suppressing noise. Specifically, matrix-based Rényi entropy~\cite{yu2019multivariate} enables direct and stable mutual information computation from embeddings without density estimation. Furthermore, the Gram matrix provides the theoretical basis for computing matrix-based Rényi entropy and also encodes pairwise similarities among samples to preserve cross-modal structure. These formulations directly support our compression and distillation losses.

\subsection{Compatibility and Extensibility of InfoCLIP} 
InfoCLIP introduces a novel distillation approach for transferring vision-language alignment, which is orthogonal to model design (e.g., the decoder). This makes it feasible and promising to integrate InfoCLIP into other advanced architectures. We apply InfoCLIP to Cat-Seg with the same decoder and training strategy, achieving superior results. Moreover, InfoCLIP offers a potential theoretical perspective for addressing the vision-language alignment problem, which may inspire future research across segmentation paradigms of foundation models (e.g., MLLM).

\end{document}